\documentclass[journal]{IEEEtran}
\usepackage{amsmath,epsfig,amsfonts,algorithmic,subfigure}
\usepackage{graphicx,cite,array,stfloats,url,longtable,float}
\usepackage{booktabs}
\usepackage{bbding}
\usepackage{multirow}
\usepackage{color, soul}
\sethlcolor{yellow}

\soulregister\cite7
\soulregister\citep7 
\soulregister\citet7 
\soulregister\ref7
\soulregister\pageref7
\soulregister\it7

\begin{document}

\title{Remote Sensing Image Translation via \\ 
Style-Based Recalibration Module and Improved Style Discriminator}

\author{Tiange Zhang, Feng Gao, Junyu Dong, Qian Du
\thanks{This work was supported in part by the National Key Research and Development Program of China under Grant 2018AAA0100602, in part by the National Natural Science Foundation of China under Grant U1706218, and in part by the Key Research and Development Program of Shandong Province under Grant 2019GHY112048. \emph{(Corresponding auther: Junyu Dong and Feng Gao)}
	
Tiange Zhang, Feng Gao, and Junyu Dong are with the School of Information Science and Engineering, Ocean University of China, Qingdao 266100, China (e-mail: zhangtiange@stu.ouc.edu.cn; gaofeng@ouc.edu.cn; dongjunyu@ouc.edu.cn;).

Qian Du is with the Department of Electrical and Computer Engineering, Mississippi State University, Starkville, MS 39762 USA (e-mail: du@ece.msstate.edu).}}

\markboth{IEEE Geoscience and Remote Sensing Letters}{}

\maketitle

\begin{abstract}

Existing remote sensing change detection methods are heavily affected by seasonal variation. Since vegetation colors are different between winter and summer, such variations are inclined to be falsely detected as changes. In this letter, we proposed an image translation method to solve the problem. A style-based recalibration module is introduced to capture seasonal features effectively. Then, a new style discriminator is designed to improve the translation performance. The discriminator can not only produce a decision for the fake or real sample, but also return a style vector according to the channel-wise correlations. Extensive experiments are conducted on season-varying dataset. The experimental results show that the proposed method can effectively perform image translation, thereby consistently improving the season-varying image change detection performance. Our codes and data are available at  \verb'https://github.com/summitgao/RSIT_SRM_ISD'
\end{abstract}


\IEEEpeerreviewmaketitle

\section{Introduction}

\IEEEPARstart{N}{owadays}, with the rapid advancement of the Earth observation program, an ever-growing number of optical remote sensing images are available. These images are widely applied in land cover change detection \cite{Turgay_Celik}, environmental analysis \cite{Bruzzone_2000_Automatic}, disaster monitoring \cite{Kahraman16}, etc. Among these applications, change detection is one of the most important techniques to monitor the changed surfaces on the Earth. In most cases, researches are mainly interested in changes caused by human activities and natural disasters. However, such changes may be difficult to be detected due to some interfering factors. For instance, vegetation colors are different between summer and winter. These changes caused by seasonal variation may yield false alarms.

Many change detection methods are affected by seasonal variation. As illustrated in Fig. 1, the change map generated by the PCAKM \cite{Turgay_Celik} contains many noisy regions. These regions are mainly vegetations whose appearances are different between winter and summer. Such variations are falsely detected as changes. Therefore, if a function can be learnt to map remote sensing images captured in summer to images captured in winter, the false alarms in change detection caused by seasonal variation can be alleviated effectively.

\begin{figure}
    \centering
    \includegraphics[width=3.4in]{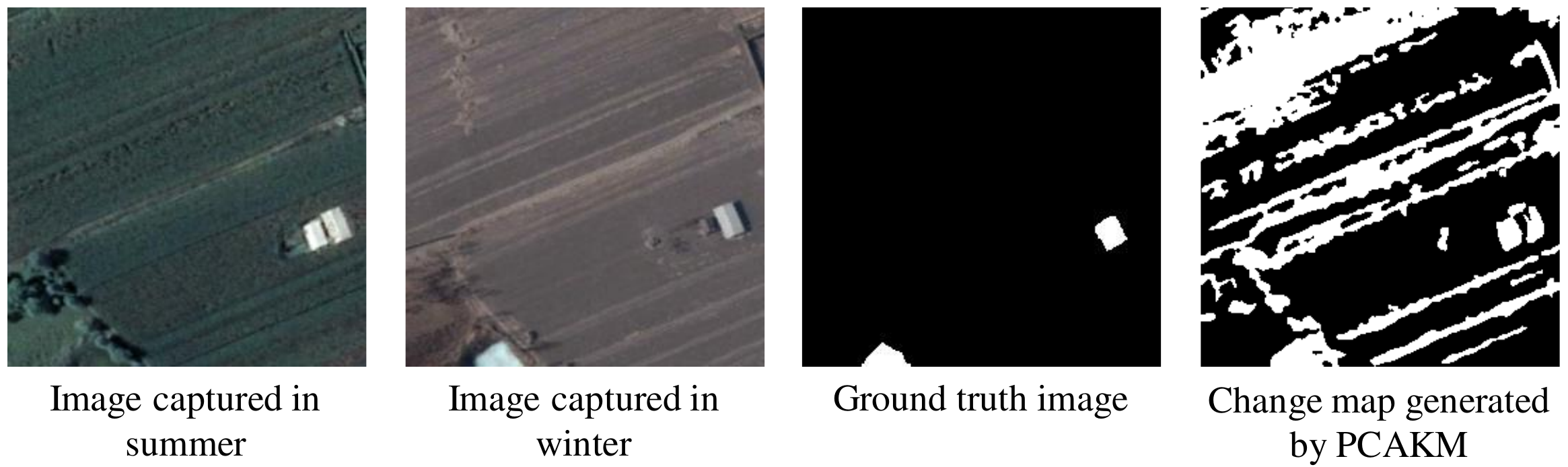}
    \caption{Illustration of the problem in change detection caused by season variation. The change map generated by PCAKM \cite{Turgay_Celik} is polluted with noisy regions. Specifically, the vegetation colors are different between summer and winter. The change detection method considers them as significant changes.}
\end{figure}

The image-to-image translation \cite{Isola_2017_CVPR} task aims to transform one representation of a scene into another. So image translation is natural to be adopted in a change detection framework to solve the aforementioned problem caused by seasonal variation. Many researchers have leveraged generative adversarial networks \cite{NIPS2014_Goodfellow} into image translation. Isola et al. \cite{Isola_2017_CVPR} used a conditional generative adversarial network (cGAN) for supervised pairwise image translation. Zhu et al. \cite{Zhu_2017_ICCV} proposed an unsupervised CycleGAN by imposing a cycle consistency loss, which is capable of learning the distinctive semantic difference between image sets from two domains and translating them correspondingly without direct pair-wise supervision.

Several attempts have been made to translate remote sensing images between different domains. Enomoto et al. \cite{Enomoto18_igarss} proposed a GAN-based image translation method for environmental monitoring. Niu et al. \cite{Niu_pcm2018} proposed an image translation method between optical and SAR data based on conditional GAN for land cover classification. Saha et al. \cite{saha20_tgrs} employed the CycleGAN to learn the transcoding between SAR and optical images, where the CycleGAN provides useful semantic features for building change detection. The above-mentioned methods are designed for the image translation task between SAR and optical images. However, image translation focusing on seasonal variation is seldom explored.

Although the tasks of SAR-to-optical translation and seasonal translation look similar, the latter is more challenging, since it focuses on transferring the seasonal features while maintaining other more important features. However, such seasonal features are difficult to be captured during the translation procedure. To tackle this issue, we consider the seasonal features of a remote sensing image as its style features, and introduce a style-based recalibration module to better capture the style features. Besides, a new style discriminator is designed to improve the translation performance during the training stage.

In this letter, the main contributions of our work can be summarized in twofold.

\begin{itemize}

\item We present an image translation method for season-varying multitemporal remote sensing images, which can effectively improve the change detection performance. To the best of our knowledge, this is the first attempt to incorporate image translation in the season-varying image change detection task.

\item To effectively capture the seasonal features, a style-based recalibration module and a new style discriminator are designed to improve the translation performance during the training phase.

\item Extensive experiments on the season-varying dataset demonstrate the superiority of the proposed method over the state-of-the-art methods. As a byproduct, we released the codes and data to facilitate other researchers.

\end{itemize}

\section{Methodology}

The architecture of the proposed method is shown in Fig. 2, where CycleGAN \cite{Zhu_2017_ICCV} is used as the backbone since it can be trained in an unsupervised way by utilizing two collections of unpaired data without labels. The model has two generators: $G_{XY}$ represents the mapping from domain $X$ (summer) to domain $Y$ (winter), while $G_{YX}$ denotes the mapping from domain $Y$ to domain $X$. Given the image collections $\left\{ x_i \right\}_{i=1}^{N_s}$ where $x_i \in X$ and $\left\{ y_j \right\}_{j=1}^{N_w}$ where $y_j \in Y$. $N_s$ denotes the number of images in the summer domain, and $N_w$ denotes the number of images in the winter domain. The discriminator $D_X$ is trained to distinguish between images $\left\{ x \right\}$ from real domain $X$ and translated images $\left\{ G_{YX}(y) \right\}$. The discriminator $D_Y$ is trained to distinguish between images $\{y\}$ from real domain $Y$ and translated images $\left\{ G_{XY}(x) \right\}$.

\begin{figure}[]
\centering
\includegraphics[width=3.2in]{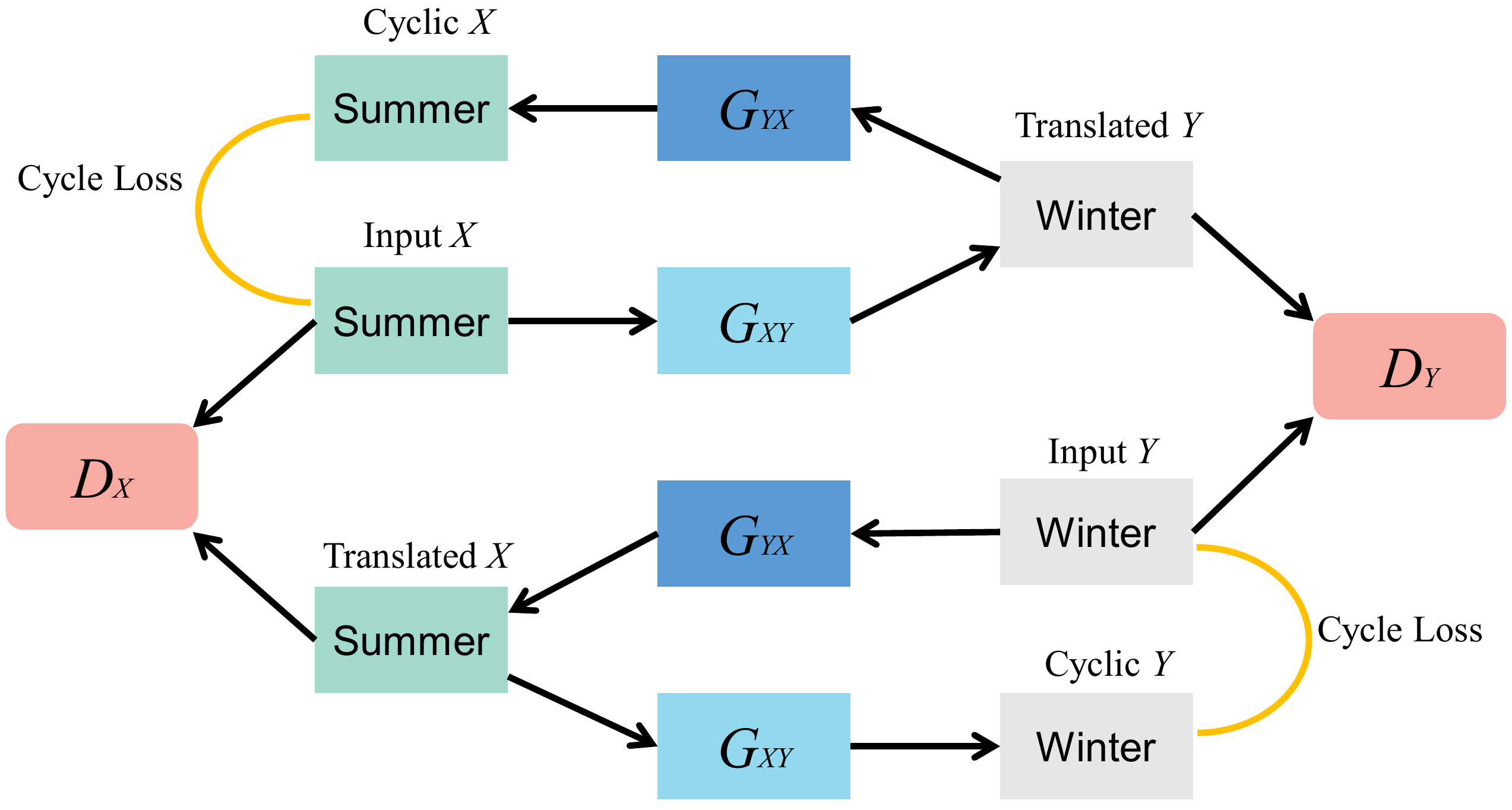}
\caption{The framework of our proposed network. Domain $X$ represents images captured in summer while domain $Y$ represents images captured in winter.}
\end{figure}

\begin{figure*}[!htb]
\centering
\includegraphics[width=4.5in]{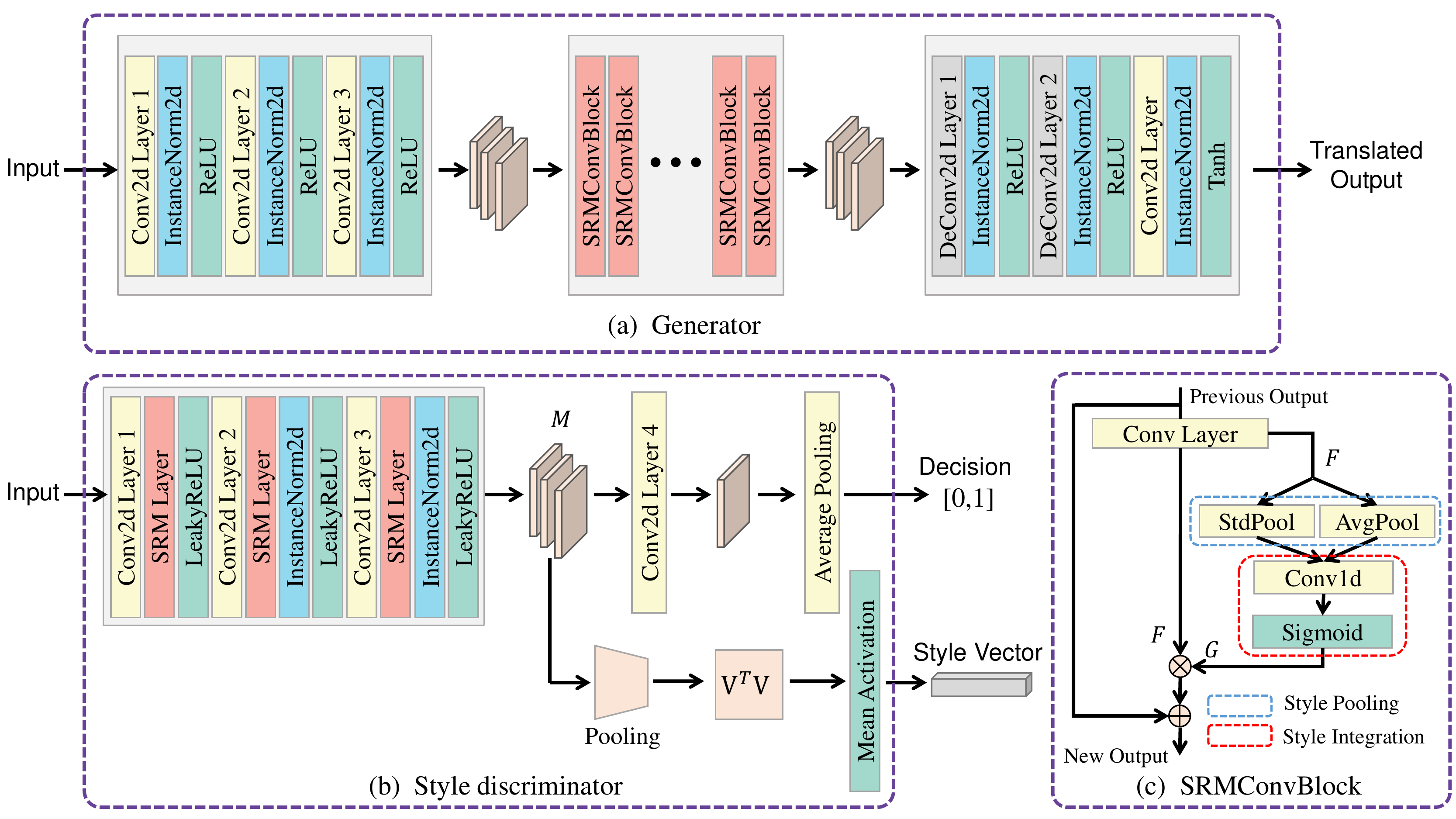}
\caption{The architecture of the basic components from our image-to-image translation network. (a) represents the Generator, which takes an image as the input and outputs the translated image. (b) represents the new Style Discriminator, which is designed to figure out whether the input is a real image or a translated one. It outputs a scalar of $[0,1]$ as the decision as well as a style vector which encode the style of the input to compute the style loss between real and translated images. (c) illustrates the SRMConvBlock utilized in both Generator and Style Discriminator.}
\end{figure*}

\subsection{The Generator with Style-Based Recalibration Module}

As shown in Fig. 3(a), the generator follows an Encoder-Transformation-Decoder structure. The encoder contains three convolutional layers to encode the image to the feature space, while the decoder utilizes two fractionally-strided convolutions and one convolutional layer to reconstruct the image from the feature space. The essential transformation module is comprised of 9 residual blocks which manipulate the feature maps to conduct the translation process.

In order to represent style features effectively in feature space, we introduce the style-based recalibration module (SRM)\cite{Lee_iccv2019} into the residual block. The SRM is comprised of two parts: \emph{style pooling} and \emph{style~integration}. The style pooling operator extracts style features from each channel by summarizing feature responses across spatial dimensions. It is followed by the style integration operator, which produces example-specific style weights by utilizing style features via channel-wise operation. Finally, the style weights recalibrate the feature maps to either emphasize or suppress their information. Note that SRMConvBlock represents the residual block which integrates SRM. Details can be found in Fig. 3 (c).

In \emph{style pooling}, the average mean and standard deviation of the feature map are selected as style features. Denote the input feature map as $F \in \mathbb{R}^{C \times H \times W}$, where $C$ is the total number of channels, and $H$, $W$ denote spatial dimensions. The style feature $T\in \mathbb{R}^{C \times 2}$ can be computed in each channel by:

\begin{eqnarray}
\mu_{c} = \frac{1}{HW} \sum\limits_{i=1}^{H}, \sum\limits_{j=1}^{W} f_{cij}
\end{eqnarray}

\begin{eqnarray}
\sigma_{c} = \sqrt{ \frac{1}{HW} \sum\limits_{i=1}^{H} \sum\limits_{j=1}^{W} (f_{cij} - \mu_{c})^2 },
\end{eqnarray}

\begin{eqnarray}
t_{c} = [\mu_{c}~,~\sigma_{c}],
\end{eqnarray}
where the style vector $t_c$ serves as a summary descriptor of the style information for $c^{th}$ channel. To be specific, $\mu_{c}$ and $\sigma_{c}$ are concatenated to form $t_{c}$.

In \emph{style integration}, the style features are converted into channel-wise style weights. Ref. \cite{Lee_iccv2019} applied a learnable channel-wise fully connected layer to achieve style integration. However, it is found that the fully connected layer has high computational burden. Therefore, in our implementation, a 1-d convolutional layer with fewer parameters but a competitive capability is used to integrate the style. Given $T \in \mathbb{R}^{C \times 2}$ from style pooling as the input, the style integration weight $G \in \mathbb{R}^{C \times 1}$ performs as the following in each channel:

\begin{equation}
g_{c} = \sigma (w * t_{c}),
\end{equation}
where $g_c$ denotes the weight for $c^{th}$ channel, $\sigma(\cdot)$ denotes the sigmoid function, $w$ denotes the convolution kernel, and * denotes convolution. Finally, the input feature $F \in \mathbb{R}^{C \times H \times W}$ is recalibrated by the channel-wise style integration weights $G \in \mathbb{R}^{C \times 1}$ to produce the output $\hat{F}\in \mathbb{R}^{C\times H \times W}$ as:
\begin{equation}
\hat{F} = G \cdot F.
\end{equation}
Here, channel-wise multiplication is employed between the style integration weight $G$ and the input feature map $F$.

\subsection{Style Discriminator}

In order to perform better image translation, we propose a new style discriminator, which not only produces a decision for the fake or real sample, but also returns a style vector according to the channel-wise correlations. By constraining the L1 loss between two style vectors of the translated image and the real target domain image, the style discriminator can distinguish the style information better.

The proposed style discriminator is illustrated in Fig. 3 (b). Given an input image, it is encoded to the feature space by adopting several convolutional layers and SRM layers. Then, the encoded feature $M$ is fed into two different modules. 

On the one hand, after convolution, $M$ is mapped into a set of overlapping image patches. Then, the decision results of each patch will be generated, and the results are averaged to a final scalar of [0, 1]. On the other hand, we consider the correlations of different channels as another kind of style representation for evaluating the style distance between translated images and target images. This correlation style representation can explore more style information across channels because in \emph{style pooling}, the style features are computed in every separate channel. First, $M \in \mathbb{R}^{C \times H \times W}$ is provided to max and average pooling layers. After each max pooling and average pooling, the aggregated feature maps are combined and fed into the next two pooling layers. Through four pooling and fusion operations, the feature map $M$ is transformed to $V \in \mathbb{R}^{C \times 1 \times 1}$. Since each channel from total number of $C$ is now represent by a single value, $V \in \mathbb{R}^{C \times 1 \times 1}$ is reshaped to $V \in \mathbb{R}^{1 \times C}$. Therefore, the correlation between different channels can be described as a matrix $V^{T}V \in \mathbb{R}^{C \times C}$ by element-wise multiplication between the transpose of $V$ and itself. Only keeping the upper triangular elements of $V^{T}V$ while setting the other elements as 0, a style vector $v$ with the length of $C \times C$ is generated after flattening  the correlation matrix. Then the L1 loss can be applied to measure the distance between two style vectors that represent translated and target images.

\subsection{Training Overflow}

As mentioned before, our goal is to learn the mapping functions $G_{XY} : X \to Y$ and $G_{YX} : Y \to X$. The objective function contains three types of terms: 1) \emph{adversarial loss} $\mathcal{L}_\textrm{GAN}$ aims to match the distribution of translated images to the data distribution in the target domain; 2) \emph{style loss} $\mathcal{L}_{S}$ calculates the differences between the translated image and the target image with two style vectors which are generated by the style discriminator $D_X$ and $D_Y$; 3) \emph{cycle consistency loss} $\mathcal{L}_{cyc}$ aims to prevent the mappings $G_{XY}$ and $G_{YX}$ from contradicting each other.

The overall objective function can be summarized as:
\begin{eqnarray}
\mathcal{L} &=& \mathcal{L}_\textrm{GAN}(G_{XY},D_Y,X,Y) + \mathcal{L}_{S}(G_{XY}(X),Y) \nonumber \\
&+& \mathcal{L}_\textrm{GAN}(G_{YX},D_X,Y,X) + \mathcal{L}_{S}(G_{YX}(Y),X) \nonumber \\
&+& \lambda\mathcal{L}_{cyc}(G_{XY},G_{YX})
\end{eqnarray}
where $\lambda$ controls the relative importance of the cycle consistency loss, and $\lambda = 10$ in the following experiments. For both $\mathcal{L}_\textrm{GAN}$, the negative log-likelihood objective is replaced with a least-squares loss, which is more stable during training and can generate higher quality samples.

Specifically, the generator $G_{XY}$ aims to minimize the following objective function:
\begin{eqnarray}
\mathcal{L}_{G_{XY}} &=& \mathcal{L}_\textrm{GAN} + \mathcal{L}_{cyc} + \mathcal{L}_{id}, ~~~~~~~~ ~~~~~~~ ~~~~~~
\end{eqnarray}
\begin{eqnarray}
\mathcal{L}_\textrm{GAN} &=& \mathbb{E}_{x\sim p_\textrm{data}(x)}[( D_{Y}(G_{XY}(x)) - 1)^2], ~~~~
\end{eqnarray}
\begin{eqnarray}
\mathcal{L}_{cyc} &=& \mathbb{E}_{x\sim p_\textrm{data}(x)}[\Vert G_{YX}(G_{XY}(x)) - x \Vert_1],
\end{eqnarray}
\begin{eqnarray}
\mathcal{L}_{id} &=& \mathbb{E}_{y\sim p_\textrm{data}(y)}[\Vert G_{XY}(y) - y \Vert_1], ~~~~~
\end{eqnarray}
where an additional loss $\mathcal{L}_{id}$ is used to regularize the generator to approximate an identity mapping when real samples of the target domain are provided as the input to the generator. The generator $G_{YX}$ operates in the same way.

For the style discriminator $D_{Y}$, it aims to minimize:
\begin{eqnarray}
\mathcal{L}_{D_{Y}} = \mathcal{L}_\textrm{GAN} + \mathcal{L}_{S}, ~~~~~~~ ~~~~~~~~ ~~~~~~~~~
\end{eqnarray}
\begin{eqnarray}
\mathcal{L}_\textrm{GAN}=\mathbb{E}_{y\sim p_\textrm{data}(y)}[(D_{Y}(y) - 1)^2] ~~~~~~~~ ~~~ \nonumber \\
+ \mathbb{E}_{x\sim p_\textrm{data}(x)}[(D_{Y}(G_{XY}(x)))^2],
\end{eqnarray}
\begin{eqnarray}
\mathcal{L}_{S}=\mathbb{E}_{x\sim p_\textrm{data}(x),y\sim p_\textrm{data}(y)}[\Vert (D_{Y}(G_{XY}(x)) - D_{Y}(y) \Vert_1],
\end{eqnarray}
where $\mathcal{L}_\textrm{GAN}$ minimizes the output decision whereas $\mathcal{L}_{S}$ regularizes the style vectors. The style discriminator $D_{X}$ is trained in the same manner.

\section{Experimental Results and Discussions}

\subsection{Datasets and Training Details}

Two sets of high resolution remote sensing images are collected for experiments from Google Earth. They are acquired at the same area from three cities in China, including Beijing, Tianjin, and Qingdao. Images in summer are captured in December 2016, and images in winter are captured in June 2019. All images are randomly cropped into 256$\times$256 pixels. We generate 869 images as the training set, and 217 images as the testing set. The proposed method is trained from scratch with a learning rate of 0.0002. We keep the same learning rate for the first 100 epochs and linearly decay the rate to zero over the next 100 epochs.

\subsection{Image-to-Image Translation Evaluation}

In order to evaluate our translation model, we employ three criteria: Inception Score (IS) \cite{Salimans_IS}, Fr\'echet Inception Distance (FID) \cite{Martin_FID}, and Kernel Inception Distance (KID) \cite{Bi2018Demystifying}. IS measures the quality and diversity of the translated images. FID compares features of real and generated images that extracted by a layer from a pre-trained Inception Network. KID computes the squared maximum mean discrepancy between the Inception feature representations of real target domain images and translated images.

\begin{table}[htpb]
\centering
\renewcommand{\arraystretch}{1.2}
\caption{Quantitative Evaluation on Translation}
\begin{tabular}{c|ccc}
\toprule
\multirow{2}{*}{Method} & \multicolumn{3}{c}{Winter to summer} \\ \cline{2-4}
& ~~ IS $\uparrow$ ~~  & ~~ FID $\downarrow$ ~~ & ~~ KID $\downarrow$ ~~ \\
\midrule
MUNIT\cite{Huang2018MUNIT} & 3.61 & 210.09 & 7.72  \\
U-GAT-IT\cite{Kim2020Ugatit} & 3.81 & 146.58 & 2.51 \\
CycleGAN\cite{Zhu_2017_ICCV} & 3.71 & 130.76 & 2.41 \\
SECycleGAN\cite{Wang2021pixel} & 3.77 & 125.08 & 1.49 \\
Ours & 3.72 & 119.75 & 1.20 \\
\bottomrule
\multirow{2}{*}{Method} & \multicolumn {3}{c}{Summer to winter} \\ \cline{2-4}
& IS $\uparrow$ & FID $\downarrow$ & KID $\downarrow$ \\
\midrule
MUNIT\cite{Huang2018MUNIT} & 3.38 & 216.41 & 8.48 \\
U-GAT-IT\cite{Kim2020Ugatit} & 3.88 & 183.26 & 5.12 \\
CycleGAN\cite{Zhu_2017_ICCV} & 3.99 & 129.02 & 1.04 \\
SECycleGAN\cite{Wang2021pixel} & 3.90 & 125.89 & 0.89 \\
Ours & 4.18 & 124.19 & 0.82 \\
\bottomrule
\end{tabular}
\label{table_1}
\end{table}

Four state-of-the-art unsupervised image-to-image translation methods including MUNIT \cite{Huang2018MUNIT}, U-GAT-IT \cite{Kim2020Ugatit}, CycleGAN \cite{Zhu_2017_ICCV}, and SECycleGAN \cite{Wang2021pixel} are selected as the baselines to evaluate the proposed method. The quantitative results are presented in Table \ref{table_1}. It shows that our proposed image-to-image translation model achieves the lowest FID and KID scores, and it is quite competitive on IS score with the CycleGAN, SECycleGAN and U-GAT-IT. These quantitative evaluation results indicate that our image translation model has learnt both mapping functions $G_{XY} : X \to Y$ and $G_{YX} : Y \to X$ well after training.

Besides quantitative evaluation, we also conduct two computational efficiency analysis on our proposed method, and the results are illustrated in Table \ref{table_2} and Table \ref{table_3}. First, Table \ref{table_2} presents the translation time of the proposed method tested on images of 256$\times$256 and 512$\times$512 pixels. It demonstrates that our method is capable of performing fast image translation. Second, Table \ref{table_3} shows the amount of parameters and GFLOPs of our image translation model compared with others. It demonstrates that the introduced style-based recalibration module and the new style discriminator can improve the translation performance only with less computational burden added.

\begin{table}
\centering
\renewcommand{\arraystretch}{1.2}
\caption{Computational Efficiency Analysis I}
\begin{tabular}{c|cc}
\toprule
\multirow{2}{*}{Mappings} & \multicolumn{2}{c}{Translation time per image (s)} \\
& ~~ 256$\times$256 ~~ & ~~ 512$\times$512 ~~ \\
\midrule
Translation (Winter to Summer) & 0.039 & 0.188 \\
Translation (Summer to Winter) & 0.040 & 0.185 \\
\bottomrule
\end{tabular}
\label{table_2}
\end{table}

\begin{table}
\centering
\renewcommand{\arraystretch}{1.2}
\caption{Computational Efficiency Analysis II}
\begin{tabular}{c|c c}
\toprule
Method &  ~~~ Params ~~~ & ~~~ GFLOPs ~~~ \\
\midrule
MUNIT\cite{Huang2018MUNIT} & 15.026 M & 77.320 \\
U-GAT-IT\cite{Kim2020Ugatit} & 10.587 M & 52.506 \\
CycleGAN\cite{Zhu_2017_ICCV} & 11.378 M & 56.832 \\
SECycleGAN\cite{Wang2021pixel} & 11.378 M & 56.832 \\
Ours & 12.634 M & 56.843 \\
\bottomrule
\end{tabular}
\label{table_3}
\end{table}

\subsection{Performance Analysis for Season-Varying Image Change Detection}

We now consider the use of the proposed method to assist in season-varying image change detection. In our approach, we first apply the method to translate multitemporal images to the same domain. Then, two well-known algorithms, PCAKM\cite{Turgay_Celik} and GETNET\cite{Wang2019getnet}, are used to obtain the change detection results. To verify the change detection performance, false alarms (FA), missed alarms (MA), overall errors (OE), and percentage correct classification (PCC) are adopted as the evaluation criteria. One typical region are selected for discussion to verify the effectiveness of the proposed image translation method.

\begin{figure}
\centering
\includegraphics[width=3.2in]{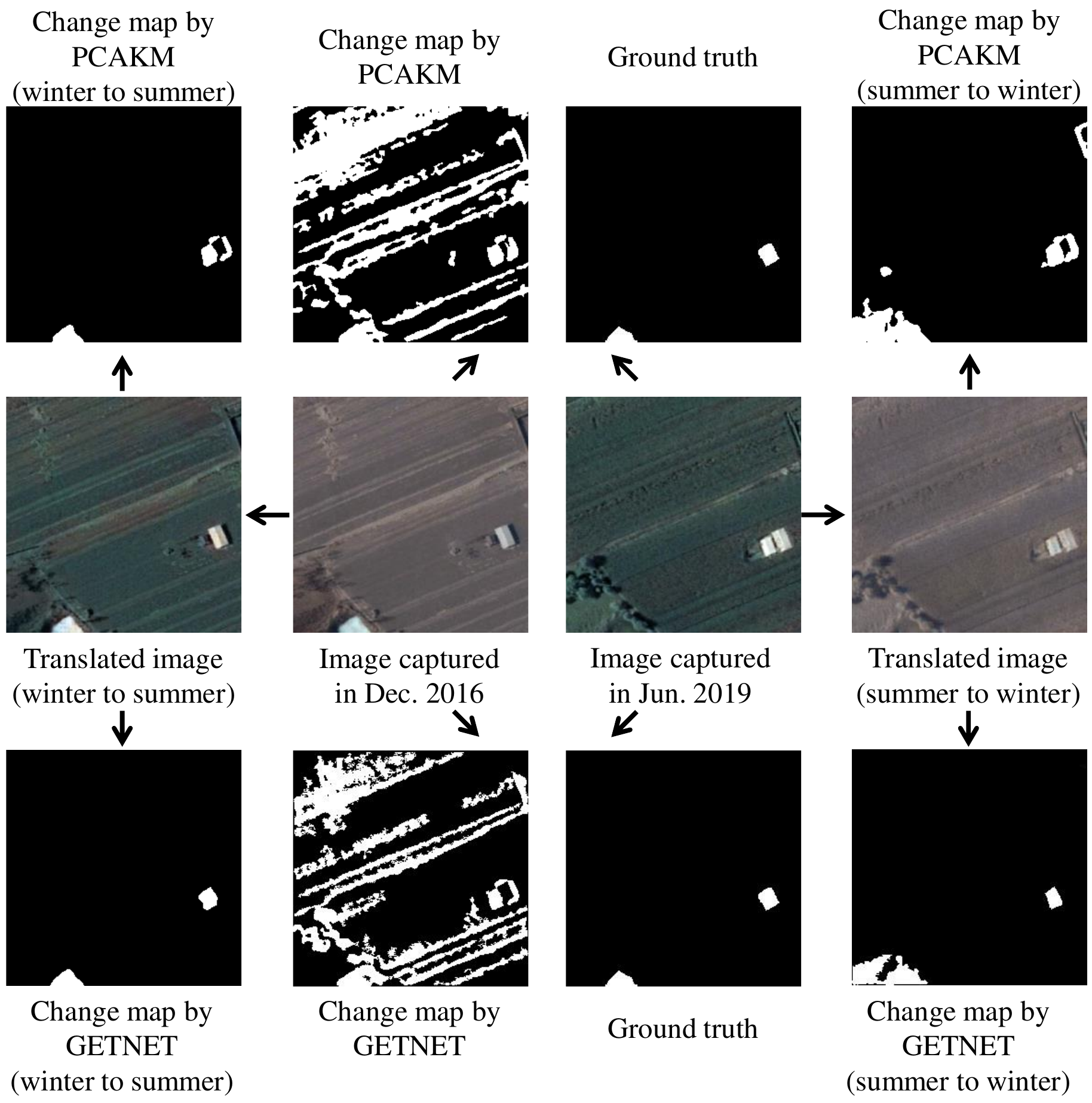}
\caption{Visualized change detection results of the selected region.}
\label{fig_change_2}
\end{figure}

\begin{table}
\centering
\renewcommand{\arraystretch}{1.2}
\caption{Change Detection Results of the selected region}
\begin{tabular}{c|cccc}
\toprule
Methods & FA & MA & OE & PCC(\%) \\
\midrule
PCAKM & 21518 & 131 & 21649 & 66.96\\
GETNET & 17663 & 144 & 17807 & 72.83\\
(Winter to Summer) + PCAKM & 390 & 146 & 536 & 99.18\\
(Summer to Winter) + PCAKM & 2682 & 26 & 2708 & 95.87\\
(Winter to Summer) + GETNET & 149 & 119 & 268 & 99.59\\
(Summer to Winter) + GETNET & 1903 & 53 & 1956 & 97.01\\
\bottomrule
\end{tabular}
\label{table_change_2}
\end{table}

As can be observed from Fig. \ref{fig_change_2}, new structures are built in this region. At first, we generate change maps directly by using PCAKM \cite{Turgay_Celik} and GETNET \cite{Wang2019getnet}. We can see that if PCAKM or GETNET is directly employed to generate the change map, many false alarms are generated since the vegetation colors are different between summer and winter. Then, Fig. \ref{fig_change_2} also demonstrates that the results generated through image translation are less noisy than the original methods. It is evident that the false alarms are greatly reduced, which means that the proposed image translation method improves the change detection performance to some extent. The corresponding quantitative results are illustrated in Table \ref{table_change_2}. By employing the seasonal translation, the PCC values are improved significantly. Moreover, the change maps generated by winter-to-summer translation achieve 99.18\% and 99.59\% for both methods, rather close to the ground truth. All these compared quantitative results reveal that the proposed scheme can produce much better change detection results generally, not only on one change detection algorithm.

\section{Conclusion}

In this letter, we present a remote sensing image translation method, which can capture seasonal features to improve the detection performance on real changes of interest. A style-based recalibration module and a new style discriminator are designed to improve the translation during the training phase. The encouraging experimental results verify the effectiveness of the proposed method and indicate the necessity of integrating the proposed translation scheme in the season-varying image change detection task. Although our method achieves good results in change detection, we believe that there is still room for improvement, such as integrating the image translation and change detection operators into an end-to-end network.

\end{document}